\documentclass[letterpaper, 10 pt, conference]{ieeeconf}  % Comment this line out if you need a4paper

\IEEEoverridecommandlockouts                              % This command is only needed if 
\usepackage{xcolor}
\usepackage{textcomp}
\usepackage{graphicx}
\usepackage{amsmath}
\let\labelindent\relax
\usepackage{enumitem}
\usepackage{soul}
\usepackage{url}

\newcommand{\textred}[1]{\textcolor{black}{#1}}
\ifx\noeditingmarks\undefined
   \newcommand{\pgwrapper}[2]{\textred{#1: #2}}
\else
   \newcommand{\pgwrapper}[2]{}
\fi
\newcommand{\textredd}[1]{\textcolor{red}{#1}}
\ifx\noeditingmarks\undefined
   \newcommand{\pgwrapperr}[2]{\textredd{#1: #2}}
\else
   \newcommand{\pgwrapperr}[2]{}
\fi

\newcommand{\name}{\mbox{MiFly}}
\newcommand{\xref}[1]{\S\ref{#1}}

\setlist[itemize]{leftmargin=0pt,itemsep=0pt,parsep=0pt, wide= 0.01pt}
\setlist[enumerate]{leftmargin=0pt, wide=0.01pt}
\newcommand{\cut}[1]{}

\newcommand{\cutt}[1]{}

\title{\LARGE \bf
3D Self-Localization of Drones using a Single Millimeter-Wave Anchor
}

\author{Maisy Lam$^{*, 1}$, Laura Dodds$^{*, 1}$, Aline Eid$^{3}$, Jimmy Hester$^{4}$,  Fadel Adib$^{1, 2}$% <-this % stops a space
\thanks{$^{*}$These authors contributed equally. $^{1}$ Massachusetts Institute of Technology, $^{2}$ Cartesian Systems, $^{3}$ University of Michigan, $^{4}$ Atheraxon}%
}

\begin{document}

\maketitle
\thispagestyle{empty}
\pagestyle{empty}

%%%%%%%%%%%%%%%%%%%%%%%%%%%%%%%%%%%%%%%%%%%%%%%%%%%%%%%%%%%%%%%%%%%%%%%%%%%%%%%%
\begin{abstract}
We present the design, implementation, and evaluation of \name, a self-localization system for autonomous drones that works across indoor and outdoor environments, including low-visibility, dark, and GPS-denied settings.

\name\ performs 6DoF self-localization by leveraging a \textit{single millimeter-wave (mmWave) anchor} in its vicinity - even if that anchor is visually occluded. MmWave signals are used in radar and 5G systems and can operate in the dark and through occlusions. \name\ introduces a new mmWave anchor design and mounts light-weight high-resolution mmWave radars on a drone. By jointly designing the localization algorithms and the novel low-power mmWave anchor hardware (including its polarization and modulation), the drone is capable of high-speed 3D localization. Furthermore, by intelligently fusing the location estimates from its mmWave radars and its IMUs, it can accurately and robustly track its 6DoF trajectory.

We implemented and evaluated \name\ on a DJI drone. We demonstrate a median localization error of 7cm and a 90\textsuperscript{th} percentile less than 15cm, even when the anchor is fully occluded (visually) from the drone.

\end{abstract}

\vspace{-0.1in}
\section{Introduction}

\vspace{-0.1in}

Drone self-localization is a classical robotics problem~\cite{classical}. It has numerous applications across outdoor and indoor environments including autonomous delivery drones~\cite{delivery}, indoor mapping~\cite{indoormap}, search-and-rescue~\cite{search}, entertainment~\cite{entertain}, and more. In outdoor environments, drones typically rely on GPS to self-localize. However, because GPS does not work indoors or when satellites are blocked by buildings, these approaches fail in indoor environments and urban canyons.
This has prompted researchers to explore other modalities for self-localization including vision/lidar (such as VIO)~\cite{VIO,VIO2}. These approaches have advanced the field of GPS-less localization, but often still struggle in dark and/or featureless environments (e.g., room with plain walls)~\cite{orb3}. Moreover, many of these vision-based approaches are more suitable for tracking changes in location rather than precise localization, which is critical for tasks like docking and delivery~\cite{dock, delivery}.\footnote{One way to overcome this is to pre-map these environments, which introduces additional overhead and new challenges.}

This paper attempts to overcome these challenges by exploring a different method for drone self-localization leveraging millimeter-wave (mmWave) anchors. Unlike visible light, mmWave signals can pass through\cutt{operate in} a variety of occlusions such as darkness, fog, snow, and many materials (e.g., cardboard, fabric, plastic, etc.). These favorable characteristics will allow a mmWave self-localization system to operate in environments where visible-light based self-localization cannot. Indeed, recent advances in the broader wireless localization field have demonstrated impressive accuracies in such settings - around 10-20~cm - using technologies like ultra-wideband~\cite{UWB_2, uwb3} (like that used in the Apple AirTags). However, to enable 3D localization, existing methods typically require many (\textit{at least three}) spatially-separated anchors within radio range for trilateration~\cite{UWB_IROS}. Thus, today's systems require deploying a large number of anchors in the environment (to ensure that at least 3 are within radio range at any point). Such requirements introduce significant infrastructure overhead, and also makes it challenging to deploy these systems for various tasks (e.g., docking at a window for delivery).
%We present \name, the first drone self-localization system capable of operating in dark, GPS-denied environments using only a single anchor.

This paper asks the following question: \textit{Can we enable accurate 6DoF drone self-localization using only a single mmWave anchor?} We present MiFly, the first drone self-localization system capable of accurate localization in dark, non-line-of-sight, and GPS-denied environments using only a single anchor. At a high level, \name\ operates by sending mmWave signals from an onboard radar to a single custom-designed anchor placed in the environment (e.g., like a sticker on a wall). Our mmWave anchor consists of a series of antennas organized in a pattern to reflect the signal back to the drone. The signal reflected back from the anchor and received by the radar allows the drone to compute its own location in the environment.

\begin{figure}
\centering
    \includegraphics[width=0.4\textwidth]{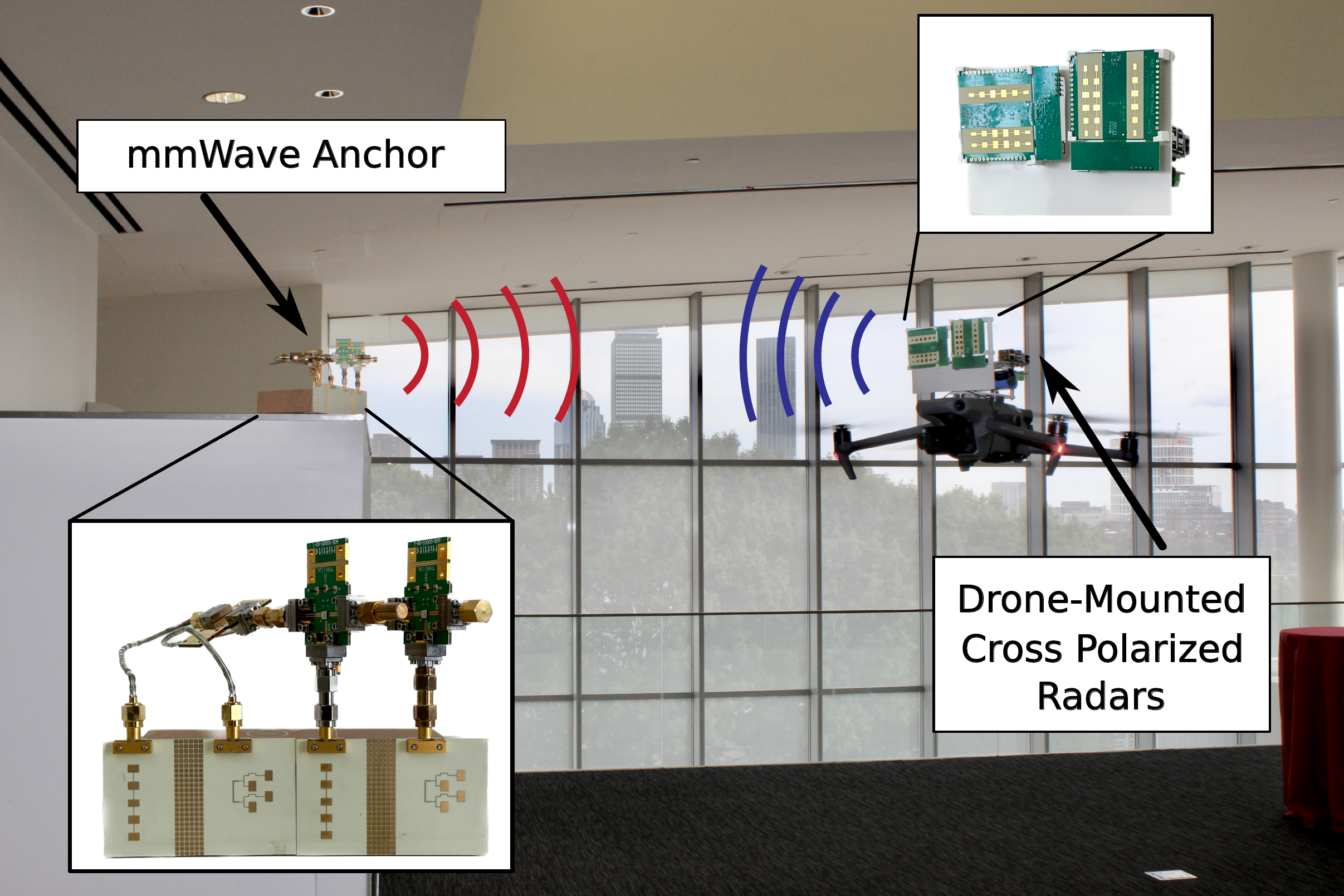}
    \caption{\footnotesize{\textbf{\name.}} \textnormal{\name\ is a self-localization system for autonomous drones. It can localize using a single mmWave anchor, allowing it to operate in low-visibility, dark, and GPS-denied settings.} }
    \label{fig:loc_err}
    \vspace{-0.25in}
\end{figure}

\begin{figure*}
\centering
        \includegraphics[width=0.8\textwidth]{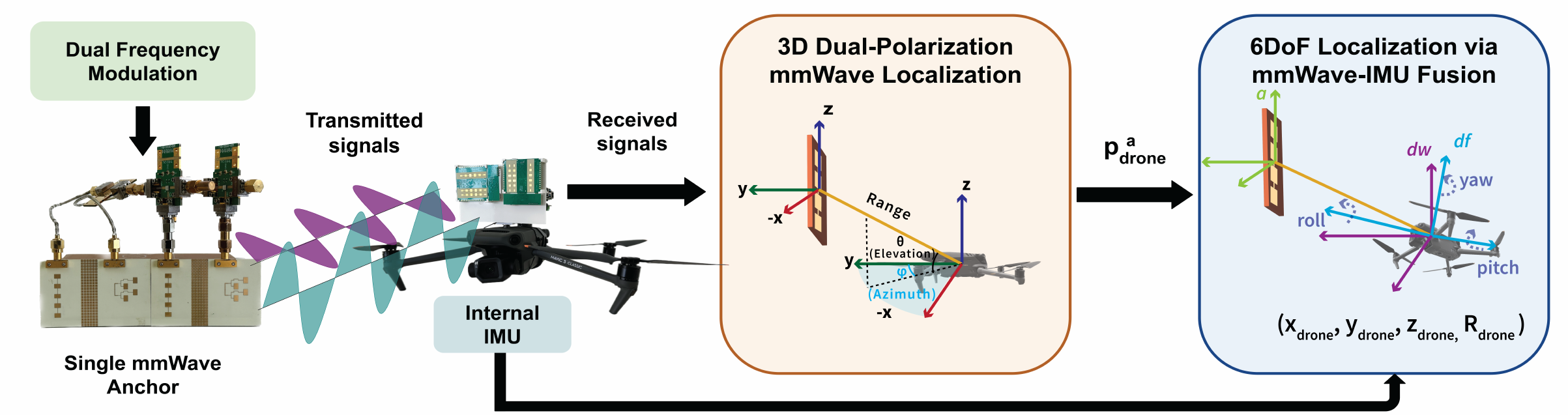}
        \vspace{-0.175in}
        \caption{\footnotesize{\textbf{System Overview.}} \textnormal{A drone with two radars measures reflected signals. Our 3D dual-polarization localization returns the range, azimuth, and elevation. Our 6DoF localization fuses the drone's internal IMU with the 3D anchor localization to compute the pose.}}
        \label{fig:system}
        \vspace{-0.25in}
\end{figure*}
To develop \name, we jointly designed a novel mmWave anchor with the drone-based self-localization algorithm. In particular, the design introduces two key contributions:
\begin{enumerate}
\item \textbf{Dual-Polarization mmWave Localization:} To enable 3D localization with a single mmWave anchor, \name\ introduces polarization into both the anchor design and localization algorithm. Polarization of electromagnetic waves is a well-known concept and has been used in optical/vision systems to filter and/or encode light. \name\ applies this concept to enable 3D localization using a mmWave anchor. In particular, unlike today's off-the-shelf mmWave radars which typically achieve high resolution in a single plane, \name's design exploits polarization to encode and filter mmWave signals across multiple planes, which enables it to perform high-resolution 3D localization. Architecturally, it leverages two orthogonal-mounted radars on the drone, and an anchor that modulates different polarizations through different frequency shifts. We describe this method in detail in~\xref{sec:single}, and how \name\ uses it for high-speed localization in~\xref{sec:high-speed}.

\item \textbf{6DoF Localization via mmWave-IMU Fusion:} While the anchor-based localization enables 3D estimates, the location has ambiguity due to the lack of rotational invariance of the angular measurement. To address this challenge, \name\ fuses the angular and range estimates obtained from the mmWave localization method with measurements obtained from the drone's onboard IMU. Not only does this approach disambiguate the 3D location, but it also enables 6DoF drone localization.
\end{enumerate}

We implement \name\ on a \textred{DJI Mavic 3 Classic} drone using two Infineon Position2Go radars. We collected over 5,500 self-localization measurements during flight. Our evaluation demonstrates a median localization error of  7cm and a 90\textsuperscript{th} percentile error less than 15cm, even when the anchor is not within the line-of-sight of the drone or is in the dark.

\section{Problem Statement}

\name\ is a 6DoF self-localization system for drones that leverages a single mmWave anchor, shown in Fig.~\ref{fig:system}. The system involves a drone equipped with two mmWave radars, each of which returns a series of measurements over time. Additionally, the drone contains a standard internal inertial measurement unit (IMU) that can provide the drone's heading at a given point in time during flight: (${roll}$, ${pitch}$, ${yaw}$). Our goal is to enable precise, minimal infrastructure, and high-speed 6DoF localization, even in GPS-denied environments and poor visibility conditions. 

To realize this vision, \name\ must satisfy the following:\cut{ requirements}

\begin{itemize}
        \item \textbf{6DoF localization}: Given the known location of an anchor, \name\ should be able to estimate the 6DoF pose of the drone 
        $(p_{drone}^a, R_{df}^a)$ with respect to the anchor.
        \item \textbf{Minimal infrastructure}: \name\ should rely on a single anchor in the environment to find its own location. This ensures minimal infrastructure, higher availability, and ease of deployment (especially in areas where placing spatially separated anchors may not be possible).
        \item \textbf{High Speed}: The self-localization must be performed at a high speed to support quick navigation reaction time. 
        \item \textbf{Self-Localization}: \name\ should be able to derive its location without needing to communicate with other devices in the environment. For example, another device separate from the drone that estimates its location and streams it to the drone would not be sufficient. This is because delays would negatively impact the drone's ability of reactive flying.
        \item \textbf{Single-Shot}: \name\ should be able to derive its location from a single measurement in time, without needing access to prior or future information. Relying on a full-time series of information would result in a slow localization scheme, which would not meet previous requirements. 
\end{itemize}

\section{Approach}
In this section, we detail the key components of \name. 

\vspace{-0.03in}
\subsection{Single-Shot, Single-Anchor, 3D mmWave Localization}\label{sec:single}

Below, we describe how \name\ achieves 3D localization, as shown in Fig.~\ref{fig:system}, by obtaining each dimension: range, azimuth ($\varphi$), then elevation ($\theta$).

\begin{figure}[t]
    \centering
    \begin{minipage}[b]{0.26\textwidth}
    \centering
            \includegraphics[width=1\textwidth,trim={0 0 0 0},clip]{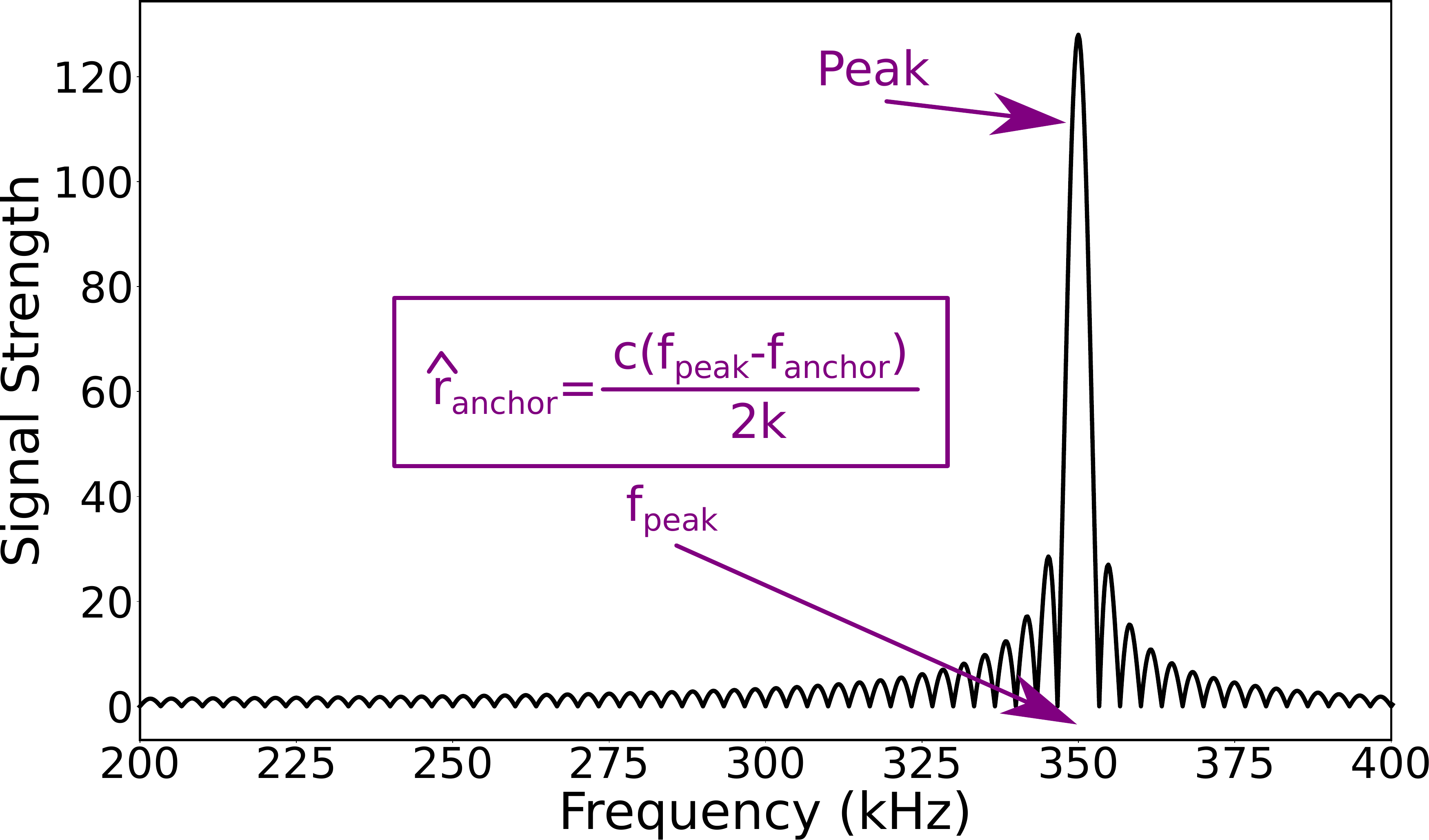}
    \vspace{-0.25in}
    \caption{\footnotesize{\textbf{Range Estimation.}} \textnormal{A simulation of the anchor's response in the FFT.}}
    \vspace{-0.2in}
    \label{fig:FFT}
    \end{minipage}
    \hspace{0.01\linewidth}
    \begin{minipage}[b]{.2\textwidth}
    \centering
            \includegraphics[width=1\textwidth]{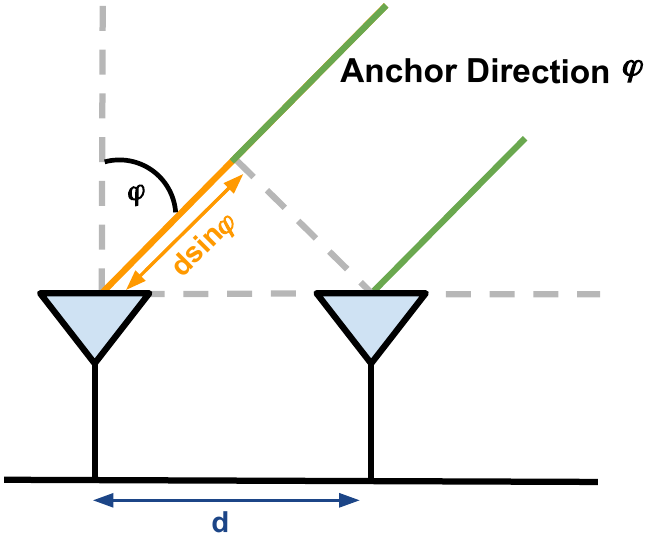}
        % \vspace{-0.25in}
        \caption{\footnotesize{\textbf{AoA.}} \textnormal{Angle-of-Arrival from drone to anchor.}}
        
        \label{fig:AoA}
        \vspace{-0.2in}
    \end{minipage}
\end{figure}

\textbf{Measuring Range.}
To measure the range, \name\ builds on past work in radar backscatter localization~\cite{rf-capture,jimmy}. In these systems, the radar transmits a chirp signal and the anchor reflects the signal back after applying a modulation (i.e., a frequency shift). Since the radar transmits a chirp signal, the frequency \cut{shift }is directly related to distance~\cite{rf-capture}. Thus, the radar can estimate the distance to the anchor using a Fourier transform (FFT) that filters around the anchor's frequency.

\begin{figure*}[t]
    \centering
    \begin{minipage}[t]{.65\textwidth}
    \centering
    \includegraphics[width=1\textwidth]{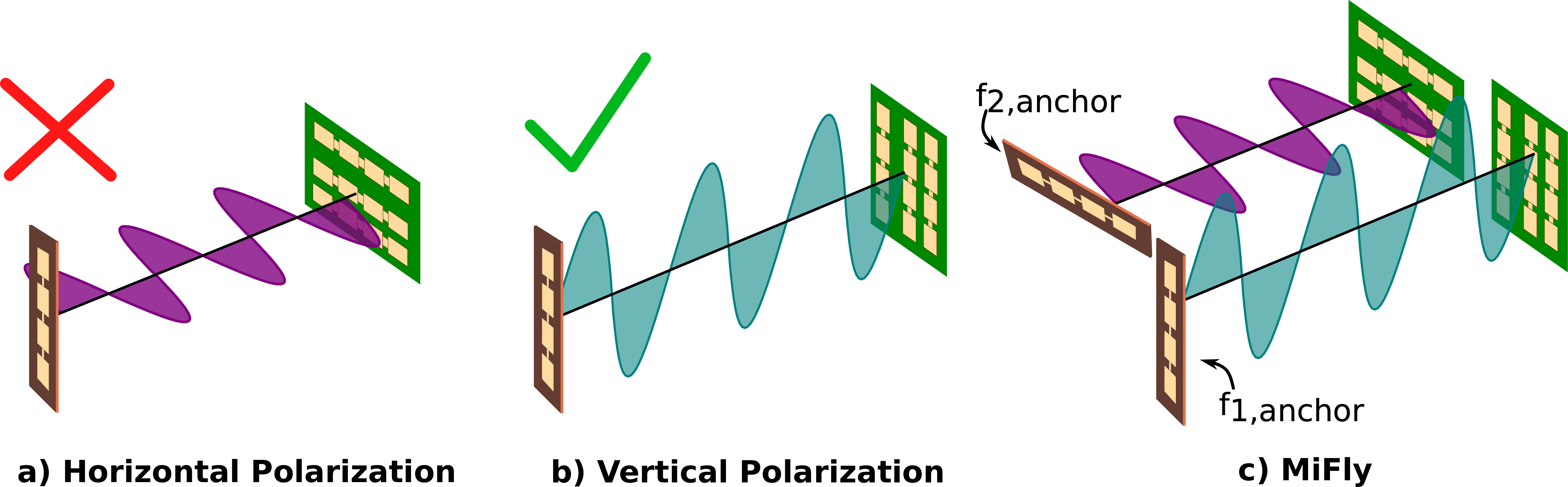}
    \caption{\footnotesize{\textbf{Polarization Diversity.} a) Horizontally polarized signals cannot be received by vertical antennas. b) Vertically polarized signals can be received by vertical antennas. c) MiFly's design leverages polarization and dual frequency modulation diversity to isolate the radar's signals.} } 
    \vspace{-0.2in}
    \label{fig:pol_div}
    % \vspace{-0.1in}
    \end{minipage}
    \hspace{0.01in}
    \begin{minipage}[t]{.32\textwidth}
    \centering
            \includegraphics[width=1\textwidth]{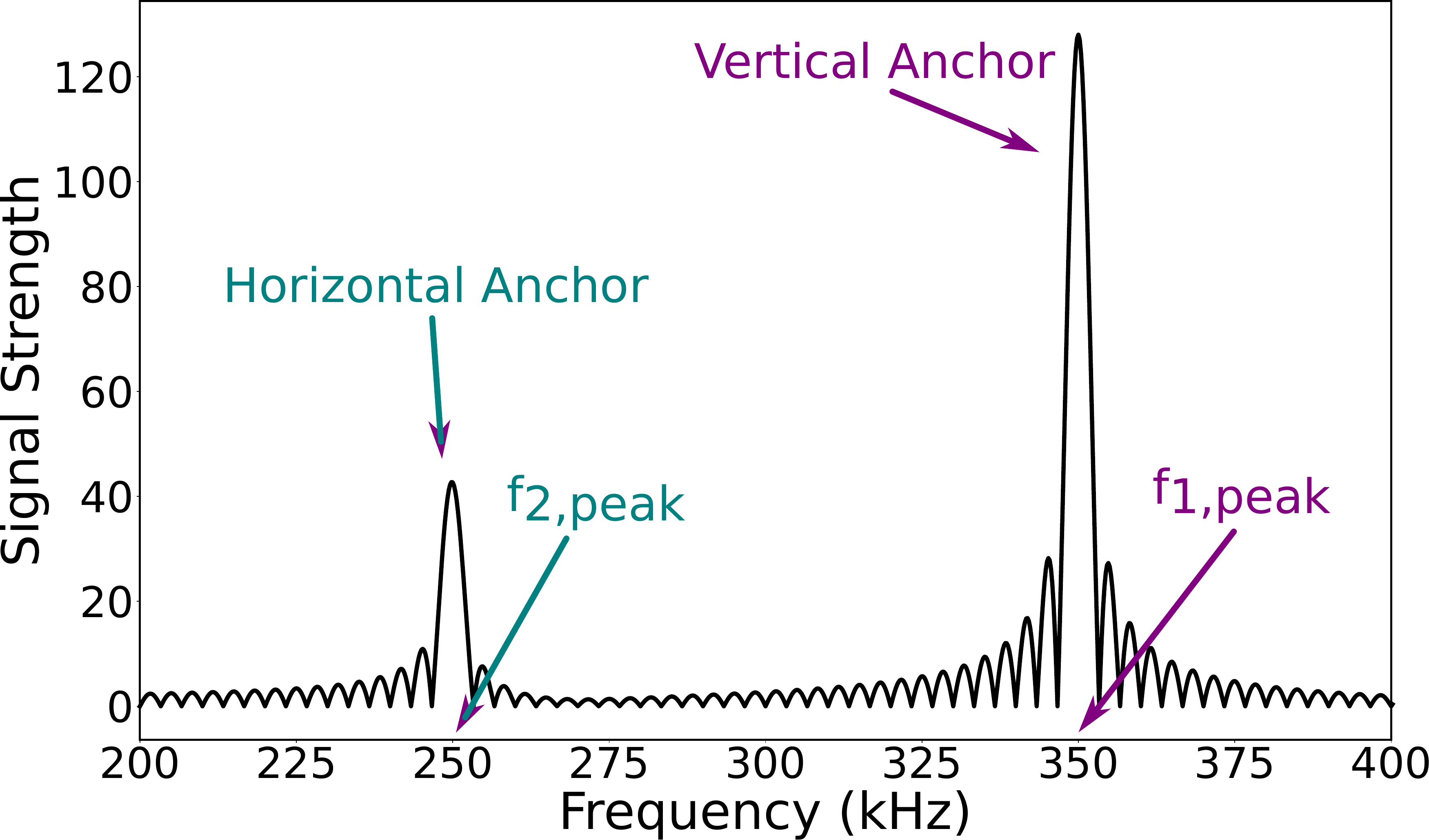}
        % \vspace{-0.1in}
        \caption{\footnotesize{\textbf{Dual Frequency Modulation.}} \textnormal{Our dual frequency modulation separates the responses from the two anchor antennas.}}
        
        \label{fig:fft2}
        % \vspace{-0.1in}
    \vspace{-0.2in}
    \end{minipage}
    % \vspace{-0.2in}
\end{figure*}

Fig.~\ref{fig:FFT} shows an illustrative example, plotting the signal received by the radar as a function of frequency.\footnote{The frequency response is shown in baseband, after chirp processing~\cite{rf-capture}.} The plot displays a peak, \textred{whose frequency} \cut{which }corresponds to the sum of the anchor's known frequency shift and the frequency caused by the travel distance. The anchor's frequency can then be used to estimate the anchor's range,$\hat{r}_{anchor}$, using the following equation~\cite{jimmy}:

\vspace{-0.05in}
\begin{equation}
\vspace{-0.05in}
\hat{r}_{anchor} =   \frac{c (f_{peak}-f_{anchor}) }{2k}
\label{eq:range}
\end{equation}

\noindent where $c$ is the speed of light, $k$ is the slope of the chirp signal, $f_{peak}$ is the frequency of the peak estimated from the FFT, $f_{anchor}$ is the known frequency of the anchor's modulation.\footnote{In practice, applying a modulation to the anchor produces two peaks within the FFT. We then use these two peaks to compute the anchor's frequency $f_{anchor}$. We refer readers to~\cite{jimmy} for more details.}

\textbf{Measuring Azimuth.} 
Next, we describe how \name\ measures the azimuth between the drone and the anchor. 
Most commodity radars can measure the angle-of-arrival (AoA), or the angle between the radar and an anchor by leveraging two or more receiver antennas in close proximity. Due to the spacing of the antennas, the anchor's signal will need to travel a small additional distance $d \sin \varphi $ to reach the second receive antenna, where $\varphi$ is the AoA or azimuth. This is shown in Fig.~\ref{fig:AoA}. Formally, the azimuth $\varphi$ is: 

\vspace{-0.05in}
\begin{equation}
% \vspace{-0.1in}
\vspace{-0.05in}
\varphi =  \sin^{-1}\left( \frac{\Delta\phi \lambda}{2 \pi d} \right)
\label{eq:aoa}
\end{equation}

\noindent where $\Delta\phi$ is the difference in phase between the two RX signals and $d$ is the physical spacing between the two antennas. We refer readers to \cite{rf-capture} for more information.

\textbf{Measuring Elevation.}
The final step to produce a full 3D location is to measure elevation. However, most commodity radars at 24GHz can only measure angle of arrival in one dimension. Instead, \name\  leverages two orthogonally-polarized radars to obtain azimuth and elevation.

For simplicity, we will first describe \name's approach assuming that the two radars operate in series (i.e., first one radar transmits and then the other). Using Eqs~\ref{eq:range}~and~\ref{eq:aoa}, \name\ computes the range and angle-of-arrival for each radar.\footnote{We average the range estimates from the two radars.} It then converts its range, azimuth ($\varphi$), and elevation ($\theta$) into a single 3D position of the anchor ($p_{anch}^{df}$) in the drone's coordinate frame during flight. Fig.~\ref{fig:system} visualizes corresponding radar and anchor (global) coordinate frames.
\vspace{-0.1in}
\begin{equation}
\vspace{-0.1in}
x = \hat{r}_{anchor}*\sin(\varphi)*cos(\theta)
\vspace{-0.03in}
\label{eq:x}
\end{equation}
\begin{equation}
\vspace{-0.1in}
y = \hat{r}_{anchor}*\cos(\varphi)*cos(\theta)
\vspace{-0.03in}
\label{eq:y}
\end{equation}
\begin{equation}
\vspace{-0.1in}
z = \hat{r}_{anchor}*\sin(\theta)
\vspace{-0.03in}
\label{eq:z}
\end{equation}
\begin{equation}
\vspace{-0.1in}
p_{anch}^{df} = (x,y,z)
%\vspace{-0.02in}
\label{eq:point}
\end{equation}

\textbf {Outlier Rejection.}
In practical implementation, the signals received by the radars are subject to noise. When noise levels are high, the probability of an incorrect localization estimation increases. To increase \name's reliability, we implement a filtering method to eliminate outliers. 

First, we define a metric to quantify the noise levels of a given measurement. We will use a metric commonly referred to as the signal-to-noise ratio, which is the ratio of the power of the received anchor's signal to the power of the received noise. We define the signal-to-noise ratio of a measurement:
\begin{equation}
    SNR = \frac{\left| \left| FFT(f_{peak}) \right| \right|^2 } {\sum_{f \in \{f_L, f_H \} - \{f_{peak}\}}   \left| \left| FFT(f) \right| \right|^2}
\end{equation}

\noindent where $FFT(f)$ is the FFT at frequency f, $\left| \left| \cdot  \right| \right|^2$ denotes the magnitude squared (power), and $f_L$, $f_H$ are the bounds of the search space over which we search for the anchor's response\footnote{We remove a small gap around $f_{peak}$ from the sum due to peak width.}

Given this SNR, we can remove any outlier measurements with a SNR below a threshold $P_{thresh}$\footnote{We additionally filter out measurements that have a large difference in estimated range between the two receiver antennas on the radar.}:
    \vspace{-0.05in}
\begin{equation}
    SNR < P_{thresh} 
    \vspace{-0.25in}
\end{equation}.

\vspace{-0.1in}
\subsection{High-Speed Localization}
\label{sec:high-speed}

In the previous section, we described how \name\ is able to produce a 3D location estimate using only a single anchor. However, so far we have assumed that the radars operate in series. This is not desirable, as it would take twice the time to produce a 3D estimate, and hinder \name's ability to support fast-flying drones. We show why simultaneous 3D estimation is challenging and describe how we custom-designed an anchor to overcome this challenge. 

The challenge in transmitting chirps from the radars simultaneously is that the chirp transmitted from one radar would reflect off the anchor and be received by the other radar, and vice versa - resulting in signal interference and preventing localization.

\name\ needs a mechanism to separate the two received signals, such that horizontal receivers only receive signals transmitted by the horizontal radar, and vertical receivers only receive those from the vertical radar.

To do so, \name\ introduces polarization as a mechanism to mitigate interference. Polarization is used in optical systems for encoding or filtering different signals and describes the way electromagnetic signals propagate through space. Fig.~\ref{fig:pol_div} shows this - a horizontally polarized signal with an electric field restricted to a horizontal plane and vertically polarized signal with an electric field lying entirely within a vertical plane. An anchor with vertical antennas will only reflect the portion of a signal that is vertical. 
Therefore, we can use two different polarizations to transmit and receive signals simultaneously, without them interfering with each other. This is known as polarization diversity.

\name\ incorporates polarization diversity into the anchor design to enable single-shot 3D localization. Since its radars are placed perpendicularly, the horizontal and vertical radar have horizontal and vertical polarizations, respectively. If we leverage an anchor design that could reflect both horizontal and vertical polarizations, (e.g., a 45\textdegree\ polarization), we would receive collided signals as described above. Instead, we design our anchor with two sets of antennas, shown in Fig.~\ref{fig:pol_div}c. The first set of antennas is horizontally polarized, and reflects only the chirps from the horizontal radar. The second set of antennas is vertically polarized and only reflects chirps from the vertical radar. Thus, we can transmit from both radars simultaneously without them interfering, allowing us to measure 3D locations in half the time.

\subsubsection{Dual-Frequency Modulation}

In theory, polarization diversity provides perfect separation between the two signals; however, in practice there is some interference and signal transmitted from the vertical radar and reflected from the vertical anchor may be detected by the horizontal receiver. \textred{This phenomenon is even more drastic in the case of a drone, because the drone's rotations during flight may cause radar rotation and misalignment, allowing each radar to receive even more signal from the opposite anchor antenna. This interference can deteriorate the performance when running radars simultaneously. } In \name, we \cut{improve the isolation between the two radars even further}\textred{overcome this challenge} through a new technique: \textit{dual-frequency modulation}.  In this technique we modulate the vertical and horizontal antennas on the anchor at different frequencies, illustrated in Fig.~\ref{fig:pol_div}c, to shift the chirps at each antenna to different frequencies. This will create two distinct peaks in the FFT, allowing us to separate the response from each anchor antenna. Formally, the peak frequencies of the signals from anchor antennas 1\&2 ($f_{1,peak}$ \& $f_{2,peak}$) can be derived from Eq.~\ref{eq:range}:

\vspace{-0.075in}
\begin{align}
    f_{1,peak} &= f_{1,anchor} + \frac{2 k \hat{r}_{anchor}}{c} \\
    f_{2,peak} &= f_{2,anchor} + \frac{2 k \hat{r}_{anchor}}{c}
\end{align}

\noindent where $f_{1,anchor}$ and $f_{2,anchor}$ are the modulation frequencies of antenna 1 and 2 on the anchor.

Since the range to both anchor antennas are the same, we can compute the difference in peak frequencies as:

\vspace{-0.05in}
\begin{equation}
\vspace{-0.05in}
    f_{1,peak} - f_{2,peak} = f_{1,anchor} - f_{2,anchor}
\end{equation}

Therefore, the peaks from different anchor antennas will be separated by the difference in modulation frequency. This allows us to isolate the reflection from a single anchor antenna by filtering the FFT around its anchor frequency, further isolating the interference between the two radars.

We show an illustrative example of our dual-frequency modulation in Fig.~\ref{fig:fft2}. Unlike before, the frequency spectrum now contains two peaks, corresponding to the two anchor antennas. The right peak is strong because the anchor's vertical antennas share the same polarization as the radar. Due to the polarization diversity, the left peak (from the anchor's horizontal antennas) is visibly weaker than the right. However, there is still a small amount of signal received due to limitations in polarization diversity, showing the importance of dual-frequency modulation for further isolation.\footnote{We note that some small amount of signal could travel from the vertical radar to the horizontal anchor to the horizontal receivers, causing interference. However, between the polarization diversity and the dual-frequency modulation, the interference is significantly reduced, such that remaining interference will not be strong enough to hinder the localization. } 
\vspace{-0.03in}
\subsection{RF-IMU Fusion}
\label{sec:IMU}
\vspace{-0.03in}
The previous sections described how we produce a high-speed 3D location from a single anchor. In this section, we describe how we fuse these angular and range estimates with the drone's IMU measurements to obtain a 6DoF pose. 

\begin{figure}
    \includegraphics[width=0.48\textwidth]{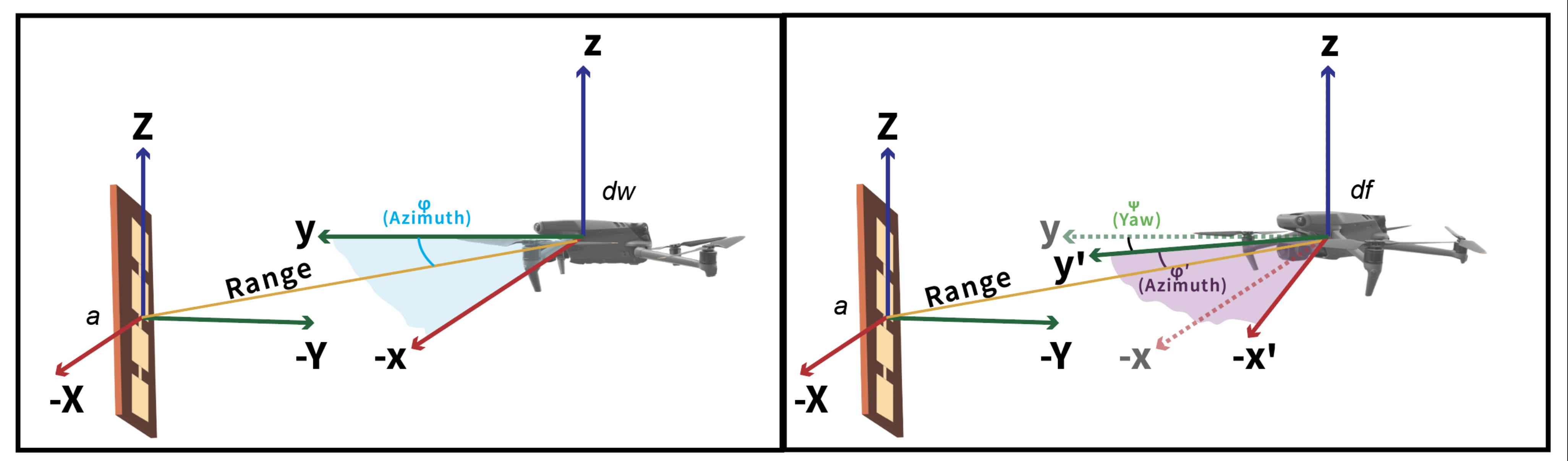}
    \vspace{-0.15in}
    \caption{\footnotesize{\textbf{Impact of Heading}} \textnormal{a) Drone directly facing anchor. b) Drone rotated about the Z axis.}}
    \label{fig:tech_heading}
    \vspace{-0.2in}
\end{figure}

A naive approach would be to directly pair the 3D location obtained via our mmWave localization techniques with the IMU detected angles. However, the anchor-based localization entangles both position and rotation information. We illustrate an example of this in Fig.~\ref{fig:tech_heading}. In Fig.~\ref{fig:tech_heading}a, the drone's radars are directed at \textred{the plane of} the anchor and the drones frames $dw$ is aligned with the anchors frame $a$. In Fig.~\ref{fig:tech_heading}b, the drone is located in the same location, but has rotated about the yaw. In this case, the azimuth detected by the radars will be different in the first case ($\varphi$ in blue) than in the second case ($\varphi$ in purple) and the mmWave localization will produce an incorrect location estimate. 

To overcome this, our system fuses the drone's internal IMU with the mmWave localization to account for the variations in heading. \cut{We first determine the 3D location of the anchor in the drone's frame during flight $p_{anch}^{df}$ using techniques detailed in \ref{sec:single} and  \ref{sec:high-speed}} \textred{Recall that the 3D location measured in \ref{sec:single} is the location of the anchor in the drone's frame during flight ($p_{anch}^{df}$ as shown in Fig.~\ref{fig:tech_heading}a)}. \textred{From the internal IMU, we can }\cut{We then }determine the rotation of drone's frame during flight \cut{$df$ }with respect to the drones world frame (${R}_{df}^{dw}$ in Fig.~\ref{fig:tech_heading}b).
\footnote{We do this by converting the IMU euler angles recorded to rotation ${R}_{df}^{dw}$} We then apply a transformation to determine the location of the anchor in the drone's world frame. The drone's world frame $dw$ at each measurement is defined as the frame possessing the translation of the drone during flight and the rotation of the anchor frame. 

\vspace{-0.1in}

\begin{equation}
\vspace{-0.05in}
p_{anch}^{dw} = {R}_{df}^{dw} *p_{anch}^{df}
\label{eq:p_anch}
\end{equation}
%\vspace{-0.05in}
In order to self localize, we must then express the location and rotation of the drone in the anchors' frame. To do this, we invert the coordinates and determine our rotation:
\begin{equation}
\vspace{-0.1in}
p_{drone}^a = -1 * p_{anch}^{dw}
\label{eq:p_drone}
\vspace{-0.03in}
\end{equation}

\begin{equation}
\vspace{-0.05in}
{R}_{df}^{a} = {R}_{dw}^a * {R}_{df}^{dw}
\label{eq:R_drone}
\end{equation}
where in our system ${R}_{dw}^{a}$ is the identity matrix following our definition of alignment of frames $dw$ and $a$.

The final 6DoF pose estimate of the drone is $(p_{drone}^a, R_{df}^a)$. 
\vspace{-0.075in}
\section{Implementation \& Evaluation}
\label{sec:evaluation}
\vspace{-0.03in}
\textbf{Physical Setup.}
We implement \name\ on a DJI Mavic 3 Classic. We mount two Infineon Position2Go 24 GHz radars, a Raspberry Pi 4 Model B and UPS power supply on the drone. We connect components using custom 3D printed parts. We connect the radars to Raspberry Pi via USB connection.  
The anchor implementation consists of two horizontally polarized and two vertically polarized antennas. The receiving antenna is followed by an HMC547ALC3 SPDT switch which routes the signal to a matched load, or feeds into an HMC342LC4 amplifier for re-emission.

\textbf{Software.} We combine the IMU information and RF measurements in post processing on an laptop running Ubuntu 22.04. Since we do not have access to the drone's software to synchronize the IMU with the Raspberry Pi, we perform a calibration to measure the time offset between the two.\footnote{We rotate the drone at the beginning of a trajectory and correlate the change in yaw from the IMU with the change in azimuth from the radars to find the time offset.} After applying the time offset, we then interpolate the IMU data to provide heading estimates at the times that the radar measurements were captured. Since the two radars are not perfectly clock synchronized, we pair the closest two measurements in time to provide a single 3D estimate.

\textbf{Evaluation Environment.} We evaluate \name\ in a typical office setting, with tables, chairs, and people moving in the background. 

\textbf{Ground Truth.} To measure the localization accuracy of our system, we use the
OptiTrack™ Motion Capture System to provide 6DoF tracking. We place tracking markers on the drone and continuously extract the 6DoF pose of the drone during flight. In the case that the drone is temporarily missed by the optitrack system, we remove that section of the trajectory. Since the ground truth is recorded on a separate computer than the drone, we use the same NTP server to provide time synchronization. We then interpolate the ground truth measurements to find the ground truth at the times that \name's measurements were recorded.

\textbf{Metric.} We measure the localization error as the 3D L2 norm difference between \name's predicted location and the ground truth location. We also measure the rotation error as the difference in degrees between the drone's estimated euler angles and the ground truth euler angles. 

\textbf{Baseline.} We implement a VIO baseline for comparison. In this baseline, we leverage an Intel® RealSense™ Tracking Camera T265. We mount the camera on the drone using 3D printed mounts. We use the cameras internal tracking algorithm to extract the 3D coordinates of the drone. 
\vspace{-0.03in}
\section{Performance Results}
\vspace{-0.03in}

\begin{figure*}[t]
    \centering
    \begin{minipage}[t]{.31\textwidth}
    \centering
        \includegraphics[width=1\textwidth]{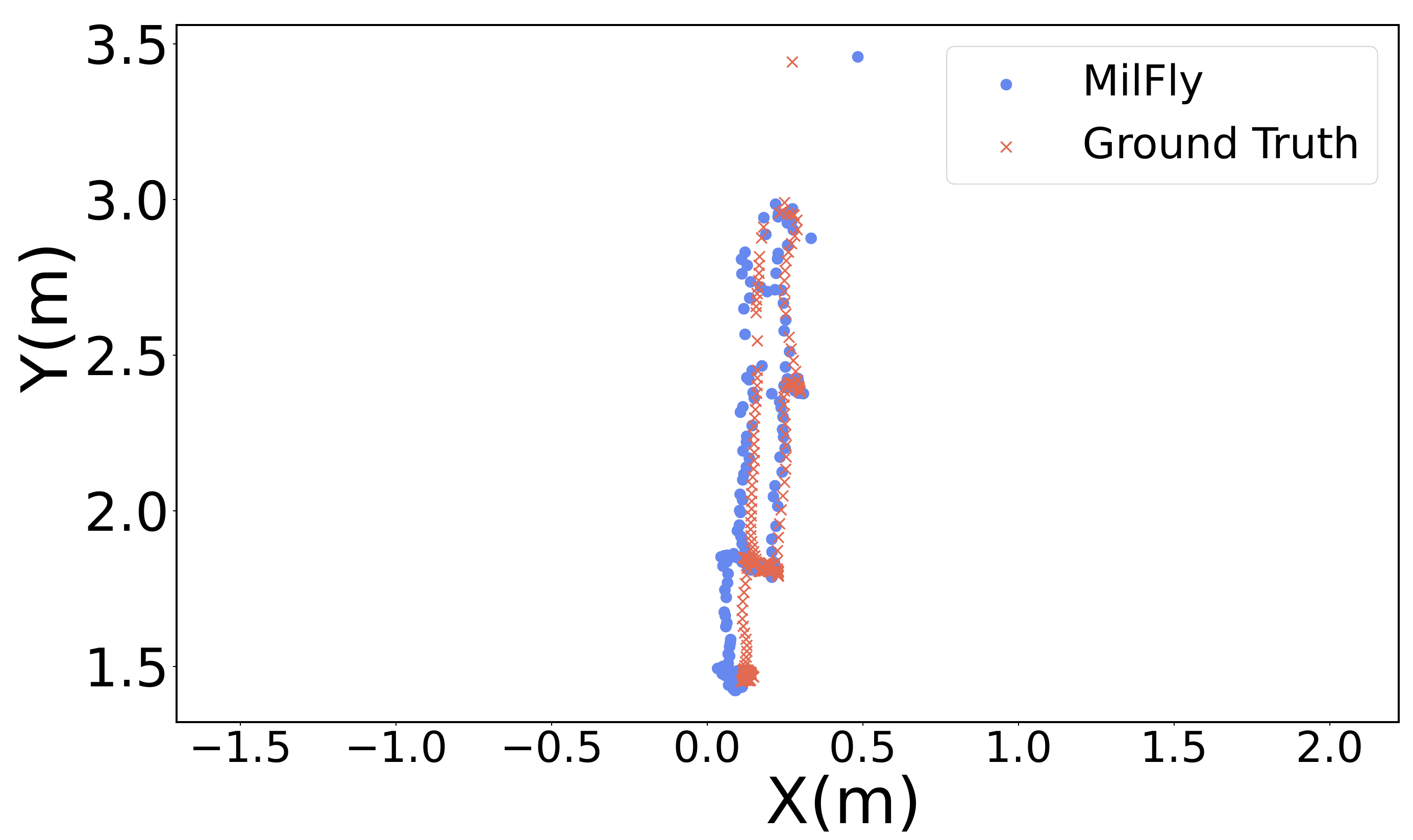}
        \vspace{-0.25in}
    \caption{\footnotesize{\textbf{Birds-Eye View.}} \textnormal{BEV (X vs Y) of an experiment for \name(blue) \& ground truth(red).} }
        \vspace{-0.05in}
    \label{fig:overhead}
    \end{minipage}
    \hspace{0.01\linewidth}
    % \hspace{-0.02\linewidth}
    \begin{minipage}[t]{.31\textwidth}
    \centering
    \includegraphics[width=1\textwidth]{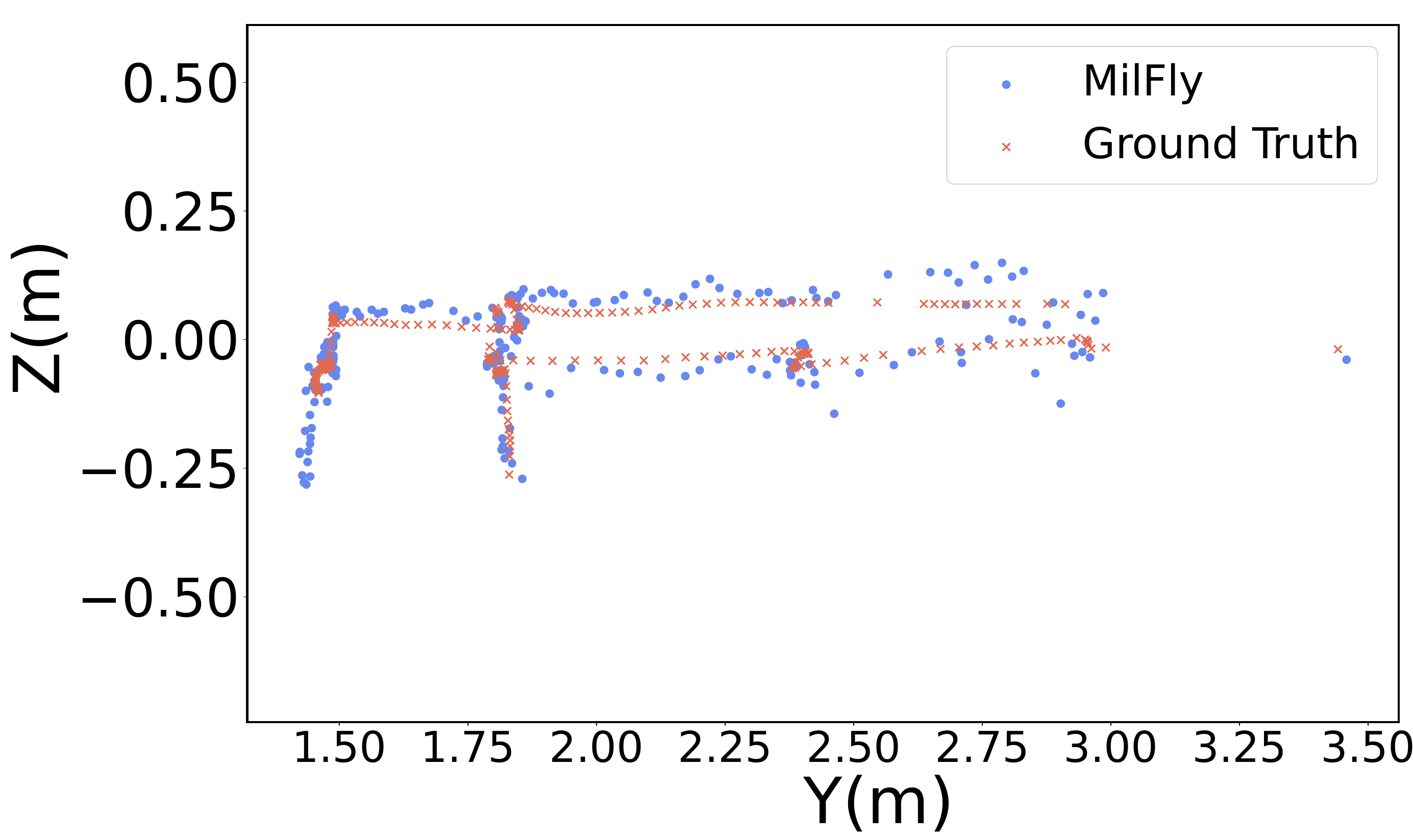}
        \vspace{-0.25in}
    \caption{\footnotesize{\textbf{Side View.}} \textnormal{Side view (Y vs Z) of an experiment for \name(blue) and ground truth(red).} }
        \vspace{-0.05in}
    \label{fig:side}
    \end{minipage}
    \hspace{0.01\linewidth}
    \centering
    \begin{minipage}[t]{.31\textwidth}
        \centering
        \includegraphics[width=1\textwidth]{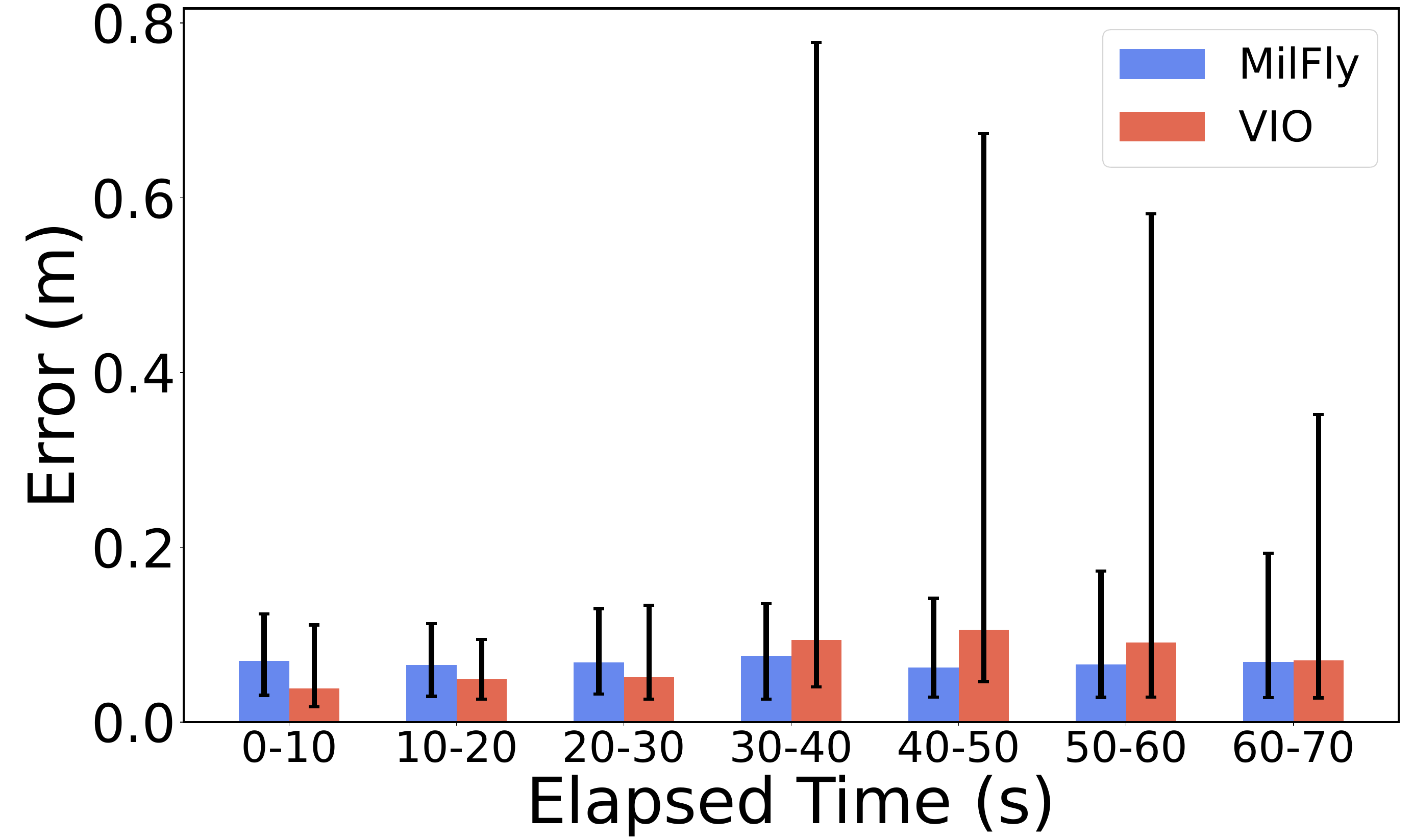}
        \vspace{-0.25in}
        \caption{\footnotesize{\textbf{Error vs. Time.}} \textnormal{3D localization error vs time for \name(blue) and VIO(red)} }
        \label{fig:time}
        \vspace{-0.05in}
    \end{minipage}
    \vspace{-0.15in}
\end{figure*}

\subsection{6DoF Error}
\label{sec:6dof_res}

To evaluate \name's overall 6DoF error, we conducted experiments where the drone was flown in the environment and \name\ continuously tracked its location. Outlier measurements were filtered using the procedure described in \ref{sec:single}. The IMU data was time aligned with mmWave measurements and used to derive the final 6DoF estimates (\ref{sec:IMU}). The ground truth was simultaneously tracked (\ref{sec:evaluation}). We collected over 5,500 localization estimates (after filtering) across 19 separate trajectories. We collected measurements in both LOS and NLOS, in light and in the dark, and while flying 1D, 2D, and 3D trajectories. We computed the 3D L2 norm error, as well as the errors in X, Y, Z, roll, pitch, and yaw.  

Table.~\ref{table:loc_err} shows the 10\textsuperscript{th}, 50\textsuperscript{th}, and 90\textsuperscript{th} percentile localization errors in X, Y, Z, roll, pitch, and yaw, as well as the 3D L2 norm. We make the following remarks:

\begin{table}[t]
\centering
% \footnotesize
\begin{tabular}{ |c|c|c|c| }
 \hline
& \multicolumn{3}{c|}{Localization \& Rotation Error} \\ \hline
 Axis &  $10^{th}$ pctl & Median & $90^{th}$ pctl  \\ \hline
 \textbf{X} & 0.8 cm & 4.7 cm & 11.1 cm \\ 
 \textbf{Y} & 0.2 cm & 1.1 cm & 3.2 cm \\
 \textbf{Z} & 0.6 cm & 3.1 cm & 9.8 cm \\ 
 \textbf{Roll} & 0.3\textdegree & 0.9\textdegree & 1.6\textdegree \\
 \textbf{Pitch} & 0.6\textdegree & 1\textdegree & 1.7\textdegree \\ 
 \textbf{Yaw} & 0.4\textdegree & 1.2\textdegree & 3.1\textdegree \\
 \hline
 \textbf{3D} & 3.0 cm & 7.0 cm & 14.5 cm \\ 
 \hline
\end{tabular}
\caption{. \textnormal{{\textbf{Localization \& Rotation Error.} The $10^{th}$, $50^{th}$, and $90^{th}$ pctl of the errors in X, Y, Z, roll, pitch, yaw, and 3D of \name.}}}
\label{table:loc_err} 
\vspace{-0.45in}
\end{table}

\begin{itemize}
    \item \name\ achieves a median L2 norm 3D error of 7cm and a 90\textsuperscript{th} percentile less than 15cm. Additionally, the 90\textsuperscript{th} percentile errors in roll, pitch, and yaw are 1.6\textdegree, 1.7\textdegree, and 3.1\textdegree, respectively. This demonstrates \name's ability to accurately self-localize using a single anchor. 
    \item Interestingly, the median error in Y is  1cm, while the median in X and Z are 4.7cm and 2.8cm respectively. This is due to the fact that error in Y is primarily determined by error in our range estimate, while error in X and Z are primarily determined by error in our azimuth and elevation estimates. In general, \name\ has slightly higher accuracy in range than AoA estimates. However, the overall error still remains low enough to enable drone self-localization. 
\end{itemize}

\vspace{-0.03in}
\subsection{Qualitative results}
\vspace{-0.01in}
Next, we show a qualitative result.  We plot \name's measurements and the interpolated ground truth measurements for a single trajectory. Fig~\ref{fig:overhead} shows a birds-eye-view of the trajectory (X vs Y in m). Fig.~\ref{fig:side} plots a side view (Y vs Z in m), for \name (blue) and ground truth(red). We note that \name\ successfully reconstructs the trajectory, with the estimated locations closely matching the ground truth.

\vspace{-0.03in}
\subsection{Comparison to VIO}
\vspace{-0.01in}

Next, we compare the performance of \name\ to that of our VIO baseline (\ref{sec:evaluation}). We define two types of environment. First, \textit{Easy} environments are well-lit, feature-rich environments favorable to VIO. Second, \textit{Hard} environments are low light and feature-less, presenting a challenge for VIO systems. We run three different trajectories in both \textit{Easy} and \textit{Hard} environments. Table~\ref{table:easy} reports the 10\textsuperscript{th}, 50\textsuperscript{th}, and 90\textsuperscript{th} percentile 3D L2 norm localization errors of VIO in m, for both easy and challenging environments. We note:

\begin{table}[t]
\centering
% \footnotesize
\begin{tabular}{ |c|c|c|c| }
 \hline
& \multicolumn{3}{c|}{VIO Localization Error (cm)} \\ \hline
 Scenario &  $10^{th}$ pctl & Median & $90^{th}$ pctl  \\ \hline
  \textbf{Easy}  & 1 & 2.1 & 4.1 \\ 
  \textbf{Hard} & 2.6 & 8.2 & 36 \\
 \hline
\end{tabular}
\caption{. \textnormal{{\textbf{VIO Performance.} The $10^{th}$, $50^{th}$, and $90^{th}$ pctl of the 3D errors of VIO in easy and hard environments.}}}
\label{table:easy} 
\vspace{-0.15in}
\end{table}
% \vspace{-0.1in}

\begin{itemize}
    \item In easy environments, VIO is able to track a trajectory with a 90\textsuperscript{th} percentile error of 4.1cm, as expected.
    \item However, when operating in challenging environments, VIO is unable to maintain its high level of accuracy, achieving a 90\textsuperscript{th} percentile of 36cm. This shows the need for \name's mmWave localization, that can operate in dark, feature-less environments (as demonstrated in Sec.~\ref{sec:6dof_res}). 
    
\end{itemize}

\vspace{-0.1in}
\section{Microbenchmarks}
\subsection{LOS vs NLOS}

\begin{table}[t]
\centering
% \footnotesize
\begin{tabular}{ |c|c|c|c| }
 \hline
& \multicolumn{3}{c|}{Localization Error (cm)} \\ \hline
 Scenario &  $10^{th}$ pctl & Median & $90^{th}$ pctl  \\ \hline
 \textbf{LOS} & 2.7 & 6.6 & 13.2 \\ 
 \textbf{NLOS} & 3.6 & 7.6 & 13.6 \\
 \hline
\end{tabular}
\caption{. \textnormal{{\textbf{Impact of Non-Line-of-Sight on Localization Accuracy.} The $10^{th}$, $50^{th}$, and $90^{th}$ pctl of the 3D errors of \name\ in LOS\&NLOS.}}}
\label{table:nlos} 
\vspace{-0.35in}
\end{table}

 In our first microbenchmark, we evaluate the impact of line-of-sight on \name's localization accuracy. We flew the drone along a specific trajectory by following markers on the floor. For each trajectory, we repeated the experiment twice, once when the anchor was within the LOS of the drone and once when the anchor was hidden from view. We ran four different trajectories for a total of 8 trajectories and over 1,200 localization measurements for both LOS and NLOS. 
 
 Table~\ref{table:nlos} shows the 10\textsuperscript{th}, 50\textsuperscript{th}, and 90\textsuperscript{th} percentiles of the 3D L2 norm localization error in cm. We note that at all percentiles, the localization error in NLOS is less than 1cm larger than in LOS. This demonstrates that \name\ is able to successfully self-localize even in the presence of occlusions.
 %since mmWave signals can traverse through many materials.

\vspace{-0.03in}
 \subsection{Error vs Time}

 Next, we investigate the error of VIO and \name\ over time. We run three different trajectories for both \name\ and VIO. For this microbenchmark, we run the trajectories in \textit{Hard} environments (e.g., darker \& feature-less). 

Fig.~\ref{fig:time} plots the median error vs elapsed time since the start of the experiment for \name(blue) and VIO(red). The error bars denote 10\textsuperscript{th} and 90\textsuperscript{th} percentiles. We note:
\begin{itemize}
\item The 90\textsuperscript{th} percentile error for \name\ is 12cm, 11cm, and 13cm for the first three time bins. The 90\textsuperscript{th} percentile error for VIO is 11cm, 9cm, and 13cm for the first three time bins. This shows that at the beginning of an experiment, the performance of VIO and \name\ are comparable. 
\item Starting at 30-40 seconds, the 90\textsuperscript{th} percentile error of VIO is 77cm, while the 90\textsuperscript{th} percentile error of \name\ is only 14cm. A similar trend continues until the end of the trials. This shows that while VIO is subject to large failures in challenging environments, \name\ is able to maintain a consistent localization performance for the entire trajectory. 
\end{itemize}

\vspace{-0.075in}
\section{Related Works}
\vspace{-0.05in}
Past work has explored different methods for drone self-localization including vision, Lidar, IMUs, wireless, and sensor fusion~\cite{lidar_drones,graph-slam,accurate_zero_cost, Steph}. The most relevant to \name\ is past work that leverages radio frequency (RF) signals for drone localization, since these can operate in dark and GPS-denied environments.
Past work in this space falls in two categories. The first relies on radars that track the drone from a distance; these are similar to airplane radars and are used to localize the drone from a vantage point, but are not suitable for drone self-localization \cite{mmwave_drone, MMHAWK,pursuin_drones_mmwave}. The second category relies on placing wireless anchors in the environment and localizing with respect to these anchors. Researchers have investigated using different wireless anchors including WiFi \cite{wifi-drone}, and Ultra-wideband \cite{UWB_IROS, UWB_2, prabal}. Unlike \name, which requires a single anchor, these systems typically require deploying at least 3 anchors in the environment to enable 3D localization. While some have explored the potential to localize using a single anchor, they have been limited to 2D and often cannot perform such localization in a single shot \cite{deepak_nsdi}. Recent proposals have made initial efforts to achieve single-shot localization with a single anchor~\cite{wifi-drone}, but they suffer from large errors (1~m median) and require high-power WiFi routers as anchors; \name\ shares the goals of these systems but achieves 10x higher accuracy and does so using low-power backscatter anchors. 

Finally, \name\ is related to past work on mmWave backscatter localization \cite{Hawkeye,Millimetro,jimmy}. Similar to past work on WiFi/UWB localization, these past systems also require deploying multiple anchors in the environment to enable 3D localization. \name\ builds on this rich body of literature and is the first two enable 3D localization using a single mmWave backscatter anchor (and recover 6DoF by fusing with IMUs).

\vspace{-0.03in}
\section{Conclusion}
\vspace{-0.03in}
We present \name, a self-localization system for autonomous drones using only a single millimeter-Wave anchor. \name\ leverages a novel dual polarized, dual modulated mmWave anchor and mmWave-IMU Fusion self-localization algorithm to ultimately achieve precise, high speed 6D localization. Our experiments verify that \name\ can robustly self-localize in GPS denied environments, adverse lighting conditions, and in non-line-of-sight to the anchor. For future work, we aim to incorporate our self-localization into the navigation of the drone and enable autonomous flight for applications such as docking, delivery and discovery. 

\bibliographystyle{plain}
\bibliography{ref}

\begin{thebibliography}{10}

\bibitem{rf-capture}
Fadel Adib, Chen-Yu Hsu, Hongzi Mao, Dina Katabi, and Fr\'{e}do Durand.
\newblock Capturing the human figure through a wall.
\newblock {\em ACM Trans. Graph.}, 34(6), nov 2015.

\bibitem{delivery}
Robin~Riedel Andrea~Cornell, Sarina~Mahan.
\newblock Commercial drone deliveries are demonstrating continued momentum in 2023, 2023.

\bibitem{Hawkeye}
Kang~Min Bae, Hankyeol Moon, Sung-Min Sohn, and Song~Min Kim.
\newblock Hawkeye: Hectometer-range subcentimeter localization for large-scale mmwave backscatter.
\newblock In {\em Proceedings of the 21st Annual International Conference on Mobile Systems, Applications and Services}, MobiSys '23, page 303–316, New York, NY, USA, 2023. Association for Computing Machinery.

\bibitem{orb3}
Carlos Campos, Richard Elvira, Juan J.~Gómez Rodríguez, José~M. M.~Montiel, and Juan D.~Tardós.
\newblock Orb-slam3: An accurate open-source library for visual, visual–inertial, and multimap slam.
\newblock {\em IEEE Transactions on Robotics}, 37(6):1874--1890, 2021.

\bibitem{wifi-drone}
Guoxuan Chi, Zheng Yang, Jingao Xu, Chenshu Wu, Jialin Zhang, Jianzhe Liang, and Yunhao Liu.
\newblock Wi-drone: Wi-fi-based 6-dof tracking for indoor drone flight control.
\newblock MobiSys '22, page 56–68, New York, NY, USA, 2022. Association for Computing Machinery.

\bibitem{UWB_2}
St\'{e}phane D'Alu, Oana Iova, Olivier Simonin, and Herv\'{e} Rivano.
\newblock Demo: In-flight localisation of micro-uavs using ultra-wide band.
\newblock In {\em Proceedings of the 2020 International Conference on Embedded Wireless Systems and Networks}, EWSN '20, page 186–188, USA, 2020. Junction Publishing.

\bibitem{pursuin_drones_mmwave}
Sedat Dogru and Lino Marques.
\newblock Pursuing drones with drones using millimeter wave radar.
\newblock {\em IEEE Robotics and Automation Letters}, 5(3):4156--4163, 2020.

\bibitem{dock}
Carlo~Giorgio Grlj, Nino Krznar, and Marko Pranji{\'c}.
\newblock A decade of uav docking stations: A brief overview of mobile and fixed landing platforms.
\newblock {\em Drones}, 6(1):17, 2022.

\bibitem{classical}
Abhishek Gupta and Xavier Fernando.
\newblock Simultaneous localization and mapping (slam) and data fusion in unmanned aerial vehicles: Recent advances and challenges.
\newblock {\em Drones}, 6(4):85, 2022.

\bibitem{Steph}
Ninad Jadhav, Weiying Wang, Diana Zhang, Swarun Kumar, and Stephanie Gil.
\newblock Toolbox release: A wifi-based relative bearing sensor for robotics.
\newblock {\em arXiv preprint arXiv:2109.12205}, 2021.

\bibitem{MMHAWK}
{Jia Zhang, Xin Na, Rui Xi, Yimiao Sun, Yuan He}.
\newblock {\em arXiv preprint arXiv:2308.06479}, 2023.

\bibitem{accurate_zero_cost}
Swarun Kumar, Stephanie Gil, Dina Katabi, and Daniela Rus.
\newblock Accurate indoor localization with zero start-up cost.
\newblock In {\em Proceedings of the 20th Annual International Conference on Mobile Computing and Networking}, MobiCom '14, page 483–494. Association for Computing Machinery, 2014.

\bibitem{entertain}
Michael J. de la~Merced Lauren~Hirsch.
\newblock Fireworks have a new competitor: Drones, 2023.

\bibitem{jimmy}
Charles~A. Lynch, Ajibayo~O. Adeyeye, J.~G.D. Hester, and Manos~M. Tentzeris.
\newblock When a single chip becomes the rfid reader: An ultra-low-cost 60 ghz reader and mmid system for ultra-accurate 2d microlocalization.
\newblock In {\em 2021 IEEE International Conference on RFID (RFID)}.

\bibitem{graph-slam}
A.~Moura, J.~Antunes, A.~Dias, A.~Martins, and J.~Almeida.
\newblock Graph-slam approach for indoor uav localization in warehouse logistics applications.
\newblock In {\em 2021 IEEE International Conference on Autonomous Robot Systems and Competitions (ICARSC)}, pages 4--11, 2021.

\bibitem{prabal}
Pat Pannuto, Benjamin Kempke, and Prabal Dutta.
\newblock Slocalization: Sub-uw ultra wideband backscatter localization.
\newblock IPSN '18, page 242–253. IEEE Press, 2018.

\bibitem{lidar_drones}
Jie Qian, Kaiqi Chen, Qinying Chen, Yanhong Yang, Jianhua Zhang, and Shengyong Chen.
\newblock Robust visual-lidar simultaneous localization and mapping system for uav.
\newblock {\em IEEE Geoscience and Remote Sensing Letters}, 19:1--5, 2022.

\bibitem{UWB_IROS}
Jorge~Peña Queralta, Carmen Martínez~Almansa, Fabrizio Schiano, Dario Floreano, and Tomi Westerlund.
\newblock Uwb-based system for uav localization in gnss-denied environments: Characterization and dataset.
\newblock In {\em 2020 IEEE/RSJ International Conference on Intelligent Robots and Systems (IROS)}, pages 4521--4528, 2020.

\bibitem{VIO2}
Mohammad~Fattahi Sani and Ghader Karimian.
\newblock Automatic navigation and landing of an indoor ar. drone quadrotor using aruco marker and inertial sensors.
\newblock In {\em 2017 International Conference on Computer and Drone Applications (IConDA)}, pages 102--107, 2017.

\bibitem{indoormap}
Ishveena Singh.
\newblock Skydio drones can now scan indoor spaces autonomously, 2023.

\bibitem{search}
Skydio.
\newblock How drones are used for search and rescue, 2023.

\bibitem{Millimetro}
Elahe Soltanaghaei, Akarsh Prabhakara, Artur Balanuta, Matthew Anderson, Jan~M. Rabaey, Swarun Kumar, and Anthony Rowe.
\newblock Millimetro: Mmwave retro-reflective tags for accurate, long range localization.
\newblock In {\em Proceedings of the 27th Annual International Conference on Mobile Computing and Networking}, MobiCom '21, page 69–82, New York, NY, USA, 2021. Association for Computing Machinery.

\bibitem{VIO}
Amr Suleiman, Zhengdong Zhang, Luca Carlone, Sertac Karaman, and Vivienne Sze.
\newblock Navion: A 2-mw fully integrated real-time visual-inertial odometry accelerator for autonomous navigation of nano drones.
\newblock {\em IEEE Journal of Solid-State Circuits}, 54(4):1106--1119, 2019.

\bibitem{deepak_nsdi}
Deepak Vasisht, Swarun Kumar, and Dina Katabi.
\newblock Decimeter-level localization with a single wifi access point.
\newblock NSDI'16, page 165–178, USA, 2016. USENIX Association.

\bibitem{uwb3}
Leehter Yao, Yeong-Wei~Andy Wu, Lei Yao, and Zhe~Zheng Liao.
\newblock An integrated imu and uwb sensor based indoor positioning system.
\newblock In {\em 2017 International Conference on Indoor Positioning and Indoor Navigation (IPIN)}, pages 1--8, 2017.

\bibitem{mmwave_drone}
Peijun Zhao, Chris~Xiaoxuan Lu, Bing Wang, Niki Trigoni, and Andrew Markham.
\newblock 3d motion capture of an unmodified drone with single-chip millimeter wave radar.
\newblock In {\em 2021 IEEE International Conference on Robotics and Automation (ICRA)}, pages 5186--5192. IEEE, 2021.

\end{thebibliography}
\end{document}